\documentclass[]{article}
\usepackage[left=2.5cm, right=2.5cm]{geometry}
\pdfoutput=1
\usepackage[utf8]{inputenc} 
\usepackage[T1]{fontenc}    
\usepackage[hidelinks]{hyperref}       
\usepackage{url}            
\usepackage{booktabs}       
\usepackage{amsfonts}       
\usepackage{nicefrac}       
\usepackage{microtype}      
\usepackage{xcolor}
\usepackage{amsmath,amssymb}
\usepackage{subcaption}
\usepackage{bm}
\usepackage{graphicx}
\usepackage{authblk}
\usepackage{lscape}
\usepackage{multirow}
\usepackage{natbib}
\usepackage[font=small,labelfont=bf]{caption}

\title{Epigenomic language models powered by Cerebras}

\author[1*]{Meredith V. Trotter}
\author[1]{Cuong Q. Nguyen}
\author[1*]{Stephen Young}
\author[1*]{Rob T. Woodruff}
\author[1]{Kim M. Branson}
\affil[1]{Artificial Intelligence and Machine Learning, GlaxoSmithKline}
\affil[*]{\small \texttt{\{meredith.v.trotter, stephen.r.young, rob.x.woodruff\}@gsk.com}}

\date{}

\begin{document}

\maketitle

\begin{abstract}
  
Large scale self-supervised pre-training of Transformer language models has advanced the field of Natural Language Processing and shown promise in cross-application to the biological ‘languages’ of proteins and DNA. Learning effective representations of DNA sequences using large genomic sequence corpuses may accelerate the development of models of gene regulation and function through transfer learning. However, to accurately model cell type-specific gene regulation and function, it is necessary to consider not only the information contained in DNA nucleotide sequences, which is mostly invariant between cell types, but also how the local chemical and structural ‘epigenetic state’ of chromosomes varies between cell types. Here, we introduce a Bidirectional Encoder Representations from Transformers (BERT) model that learns representations based on both DNA sequence and paired epigenetic state inputs, which we call Epigenomic BERT (or EBERT). We pre-train EBERT with a masked language model objective across the entire human genome and across 127 cell types. Training this complex model with a previously prohibitively large dataset was made possible for the first time by a partnership with Cerebras Systems, whose CS-1 system powered all pre-training experiments. We show EBERT’s transfer learning potential by demonstrating strong performance on a cell type-specific transcription factor binding prediction task.  Our fine-tuned model exceeds state of the art performance on 4 of 13 evaluation datasets from ENCODE-DREAM benchmarks and earns an overall rank of 3rd on the challenge leaderboard.  We explore how the inclusion of epigenetic data and task-specific feature augmentation impact transfer learning performance. 

 \end{abstract}

\section{Introduction}

Recent work has shown promise in building language representation models from DNA sequence \citep{dnabert, cornbert} and protein sequence \citep{bepler_berger}. Both intuition and evidence \citep{dnabert, cornbert, bepler_berger, bigbird} suggest that genomic language models, having learned the underlying structure of the genome in a self-supervised manner, can be fine-tuned to transfer-learn supervised biological classification tasks more quickly and with improved generality over randomly initialized models. However, we know that DNA sequence alone may contain insufficient information to model many important genomic processes, such as transcription factor binding, gene expression and splicing. All cells in the body share a common genome, and the phenotypic differentiation we see in different cell types is controlled, at least in part, by epigenetic differences.

Genome-wide maps of chromatin accessibility, histone modifications, transcription factor binding, methylation patterns, and three-dimensional genome structure provide the current best chromatin state signals to understand variation in gene regulation across cell types, and measuring/quantifying epigenetic information is an active area of research \citep{encode, roadmap}. We refer to the combination of DNA sequence paired with cell type specific epigenetic information as an epigenome. This work introduces the first genomic language model trained with both genetic and epigenetic information as inputs, which we call Epigenomic BERT or EBERT. For comparison purposes we also trained the model with DNA inputs alone, which we call DBERT.

We use a Bidirectional Encoder Representation from Transformers (BERT) architecture \citep{devlin} similar to the recently published DNABERT model \citep{dnabert} with the addition of epigenetic inputs. This modeling strategy yields a larger training dataset and, crucially, the ability to perform inference on unseen cell types. To add such epigenetic information to our model, we use the IDEAS36 dataset \citep{ideas36} of inferred epigenetic state for 127 different cell types as a second input alongside DNA sequence information.

IDEAS is a multiple-epigenome segmentation model whose outputs define 36 highly reproducible epigenetic `states' at nearly base-pair resolution.  The IDEAS model infers these 36 epigenetic states from co-occurrence patterns, both along the human genome and across cell types, of 14 unique chromatin marks. Each IDEAS state thus represents a combinatorial pattern of a subset of these epigenetic marks. IDEAS marks 46.6\% of the human genome as epigenomically variable -- that is, having different epigenetic states in different cell types. We hypothesize that adding this epigenomic variation to our genomic language model will improve its generalizability and performance on downstream fine-tuning tasks, particularly of cell type specific models.

The datasets required to power a multi-cell type language model like EBERT are of very large scale. For models of this type, the biggest performance gains happen when dataset size and model size scale together \citep{kaplan}. Using standard GPU hardware, training a model with a large capacity appropriate to the dataset would have taken weeks or months. Cerebras Wafer Scale Engine (WSE) within the Cerebras CS-1 system enabled us to train models of this scale in just a few days, which also made it feasible to explore hyperparameters and architectures. Here we present results for a BERT$_{BASE}$ model configuration in part to facilitate comparison to existing work (DNABERT \citep{dnabert}). 

To assess the usefulness of our epigenomic language model for transfer learning, we use EBERT (and its DNA-only sibling, DBERT) as pre-trained bases to build fine-tuned transcription factor binding site prediction models using the ENCODE-DREAM transcription factor binding within-cell type benchmark datasets (\url{https://www.synapse.org/\#!Synapse:syn6131484}).

Transcription factors (TFs) are proteins that bind to DNA based on sequence specificity and by their interaction with DNA modulate gene expression and other regulatory processes \citep{lambert}. However, DNA alone is insufficient to predict all binding, because binding also depends on DNA shape features, chromatin accessibility and structure \citep{slattery}.

TF binding prediction thus provides an excellent fine-tuning use case for our model for two reasons. First, binding is determined by DNA sequence and also by local shape and epigenetic marks. We expect the IDEAS state data to contain signal for the epigenetic marks typical of a given sequence-cell type pair, which should improve the model’s ability to predict binding over a DNA-only model. Second, the IDEAS epigenetic states are inferred from cooccurrence data, and DNase-seq (chromatin accessibility) is one of the inputs to that model. It is interesting to compare the effects of including the IDEAS inferred-state and of directly-measured DNase-seq data from ENCODE.

Our work addresses three questions:
\begin{enumerate}
\item Does the inclusion of epigenetic state information (IDEAS) in the pre-trained language model improve the power of the learned genomic embedding as judged by the TF binding prediction task?
\item Does the inclusion of tissue-specific auxiliary features in addition to the language model features improve TF binding prediction?
\item How competitive are our models relative to the current state of the art as measured by performance on the ENCODE-DREAM benchmark datasets? 
\end{enumerate}

While absolute performance on the fine-tuning task is important, our main aim is to understand relative performance differences between different variants of genomic language models. This work is intended primarily as a test of concept for the epigenomic language model, and not as an in-depth research investigation of any particular fine-tuning task.

\section{Methods}

\subsection{Pre-training}

\subsubsection{Genomic and epigenetic data}

For each cell type in the IDEAS36 dataset, each 200 base pair (bp) segment of the human genomic DNA sequence has an associated IDEAS state. These IDEAS states can also be inferred for new cell types if the 14 epigenetic marks used as input to IDEAS36 model are available. Thus, training EBERT to learn the shared language of DNA and IDEAS state with data from many different tissue types may allow the model to generalize to unseen cell types in downstream tasks.

To produce the inputs to EBERT, we begin with a 1,000 bp segment of DNA and the associated IDEAS states for that genome segment in a given cell type. We tokenize and embed the DNA and IDEAS inputs separately before feeding them into the BERT model. To quantify the effects of the epigenetic inputs on EBERT performance, we also pre-trained a BERT model using only DNA as inputs, which we refer to as DBERT. Our EBERT and DBERT models both use a similar architecture to the published DNABERT model \citep{dnabert}. We pre-trained our own DBERT in order to explore different tokenization strategies and architectures to those available in the public DNABERT repository \url{https://github.com/jerryji1993/DNABERT}.  Comparing fine-tuning performance using EBERT versus DBERT as the pre-trained base model illustrates the benefits of adding epigenetic information in the form of IDEAS state.

\subsubsection{Tokenization and embedding of DNA sequences and IDEAS states }

For DNA and IDEAS state sequences, we begin with paired 1000 bp sequences. Each is tokenized into $k$-mers with a sliding window to produce a sequence of tokens with length $L_{input}$. We pre-trained EBERTs with $k \in \{5,6,7\}$ and DBERT with $k \in \{6,7\}$. Our general findings on tokenization schemes agree with other DNA sequence embedding models, DNABERT \citep{dnabert} and EP2vec \citep{ep2vec}, that found slight increases in downstream performance with increasing $k$ up to 6, with diminishing returns as $k$ increases past 6 up to 10.  Here we show results from our best performing EBERT: $k$=7 with stride of 7, which produces a $L_{input}$ of 150 tokens.  This tokenization produces a DNA-token vocabulary of size 16,384. IDEAS state $k$-mers are pooled based on the most frequent IDEAS state in that $k$-mer, producing a matched pair of sequences with a single IDEAS state for each DNA token.

\begin{figure}
  \centering
  \includegraphics[width=0.6\linewidth]{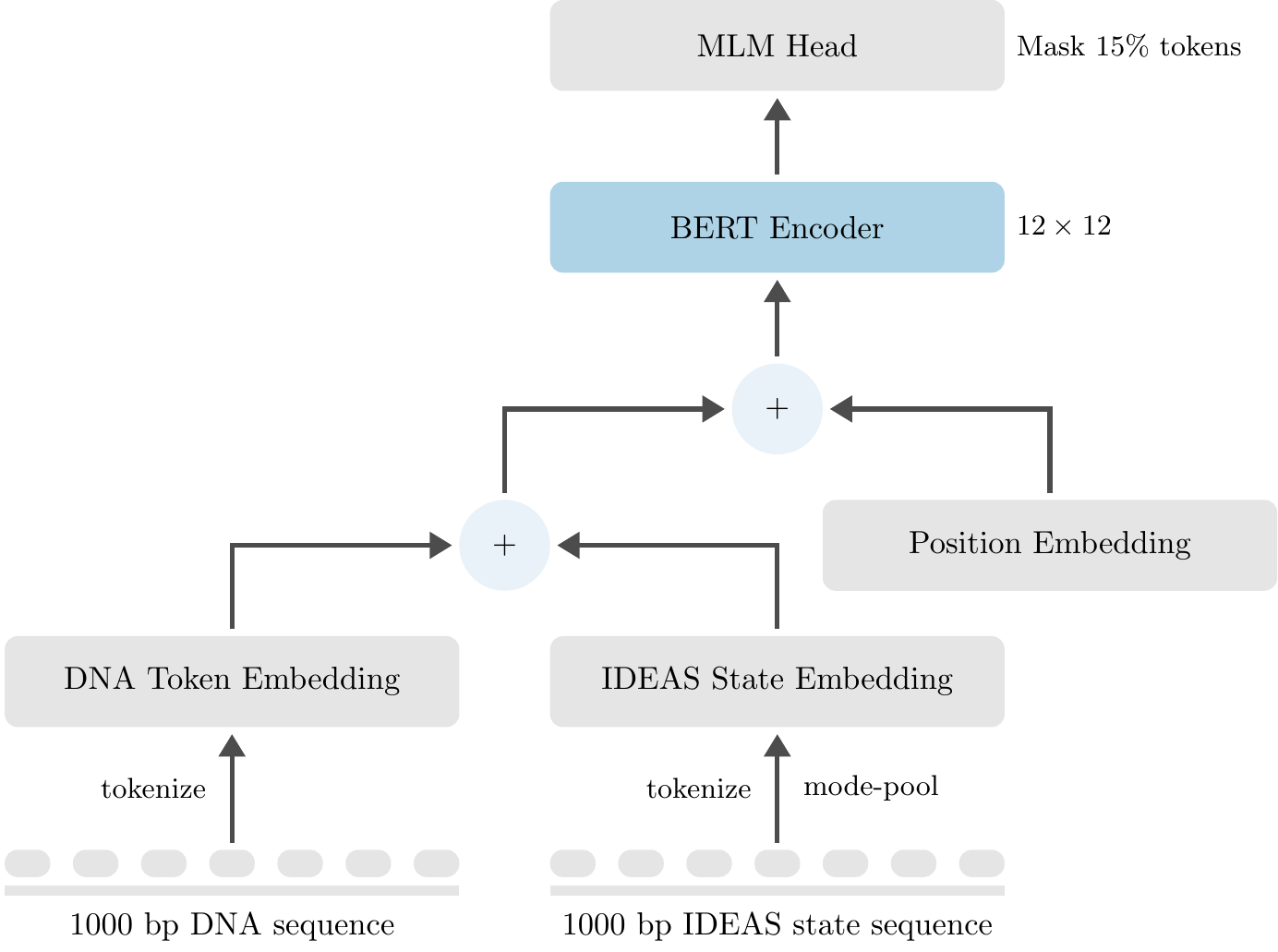}
  \caption{EBERT pre-training architecture. DBERT pre-training omitted the IDEAS state inputs and combined embedding but was otherwise identical.}
  \label{fig:arch}
\end{figure}

Tokenized DNA and IDEAS sequences initially go through separate embedding layers, which are combined before being added to a positional embedding. The resulting embedding is then the input to our central BERT encoder (Figure \ref{fig:arch}). 

\subsubsection{BERT encoder}
For our central BERT encoder we adopt a RoBERTa -style Transformer architecture \citep{roberta} with $L$ layers, where each block uses $A$ self-attention heads with hidden dimension $H$. We trained a BERT$_{BASE}$ ($L \times A \times H$ = 12 $\times$ 12 $\times$ 768) sized architecture for the encoder block. We used a paired masked-language model objective for pre-training, jointly predicting the masked tokens corresponding to the DNA and IDEAS state for a given position with a weighted cross entropy loss as expressed by

\[ \mathcal{L}_{total} = \mathcal{L}_{DNA} * \alpha +  \mathcal{L}_{IDEAS} * (1 - \alpha)\] 

where $\mathcal{L}_{total}$ denotes the total loss, $\mathcal{L}_{DNA}$ and $\mathcal{L}_{IDEAS}$ are the DNA and IDEAS losses, and $\alpha$ is the weight term. Here we show results for equal weighting ($\alpha=0.5$). For each tokenized DNA+IDEAS input, we randomly mask 15\% of the positions within the DNA sequence. Following the procedure laid out by \citet{devlin}, the tokens selected for masking are replaced with the $[MASK]$ token 80\% of the time, a random token 10\% of the time, and itself 10\% of the time. The IDEAS sequence has a resolution of 200bp, so remains the same for $200/k$ token spans.  In order to avoid information leakage from adjacent tokens, we use a modified whole-word masking on the IDEAS tokens. We randomly choose one of the masked DNA positions and mask the corresponding IDEAS token, as well as $100/k$ tokens on each side. These masked flanking tokens are masked during training but are not predicted, so do not contribute to the final loss value.

\subsection{Fine-tuning}

\subsubsection{Task-specific auxiliary features}

The top performing teams \citep{factornet, jteam, anchor} in the original ENCODE-DREAM competition all used auxiliary features in addition to primary DNA sequence as inputs to their predictive models. For example, we know chromatin accessibility information (DNase-seq data) is particularly important to TF binding, and indeed was used by all top performing models \citep{factornet, jteam, anchor}. DNA sequence uniqueness (also known as “mappability”) is closely related to the quality of sequencing data, and can improve performance by downweighting low-mappability sequences \citep{factornet}. We chose to include DNase-seq and mappability as auxiliary features in the post-BERT convolutional layers of our fine-tuning model.

Although DNase-seq data was also an input to the IDEAS states model, the IDEAS states including the most DNase-seq signal are also among the least reproducible of the states \citep{ideas36}. Additionally, the IDEAS states are inferred whereas the raw DNase-seq is specific to the TF-cell line for a given ENCODE dataset. 

DNABERT \citep{dnabert} claimed high performance on a TF binding task but did not cite performance on these specific benchmark sets. However, their finding suggests that Epigenomic BERT may not need auxiliary data in order to perform well. We compare the performance of EBERT TF binding both with and without auxiliary features to assess the relative importance of DNA+IDEAS inputs versus DNase. We compare our best performing fine-tuned models against the ENCODE-DREAM within-cell type competition leaderboard as an absolute measure of the transfer-learning potential of EBERT.

\subsubsection{Transcription factor binding model}

In the fine-tuning stage, tokenized DNA and IDEAS go through the BERT encoder without the masked language modeling head, producing an output sequence of hidden states. In the case of DBERT and EBERT fine-tuning, the sequence of hidden states (with shape $L_{input} \times H$)  feeds directly into a 2-layer 1D convolutional module followed by a 2-layer dense classification module. In the case of EBERT+, DNase-seq and mappability sequence features undergo pooling within sliding windows of size $k$ with stride $s$ to extract information matching the sequence resolution of the DNA/IDEAS tokens. These pooled auxiliary features are concatenated onto the sequence of hidden states from the BERT encoder, creating a final matrix of shape $L_{input} \times (H + 2)$. These concatenated outputs go through the convolutional and classification modules to produce the final prediction of TF binding. 

\begin{figure}[h]
  \centering
  \includegraphics[width=0.8\linewidth]{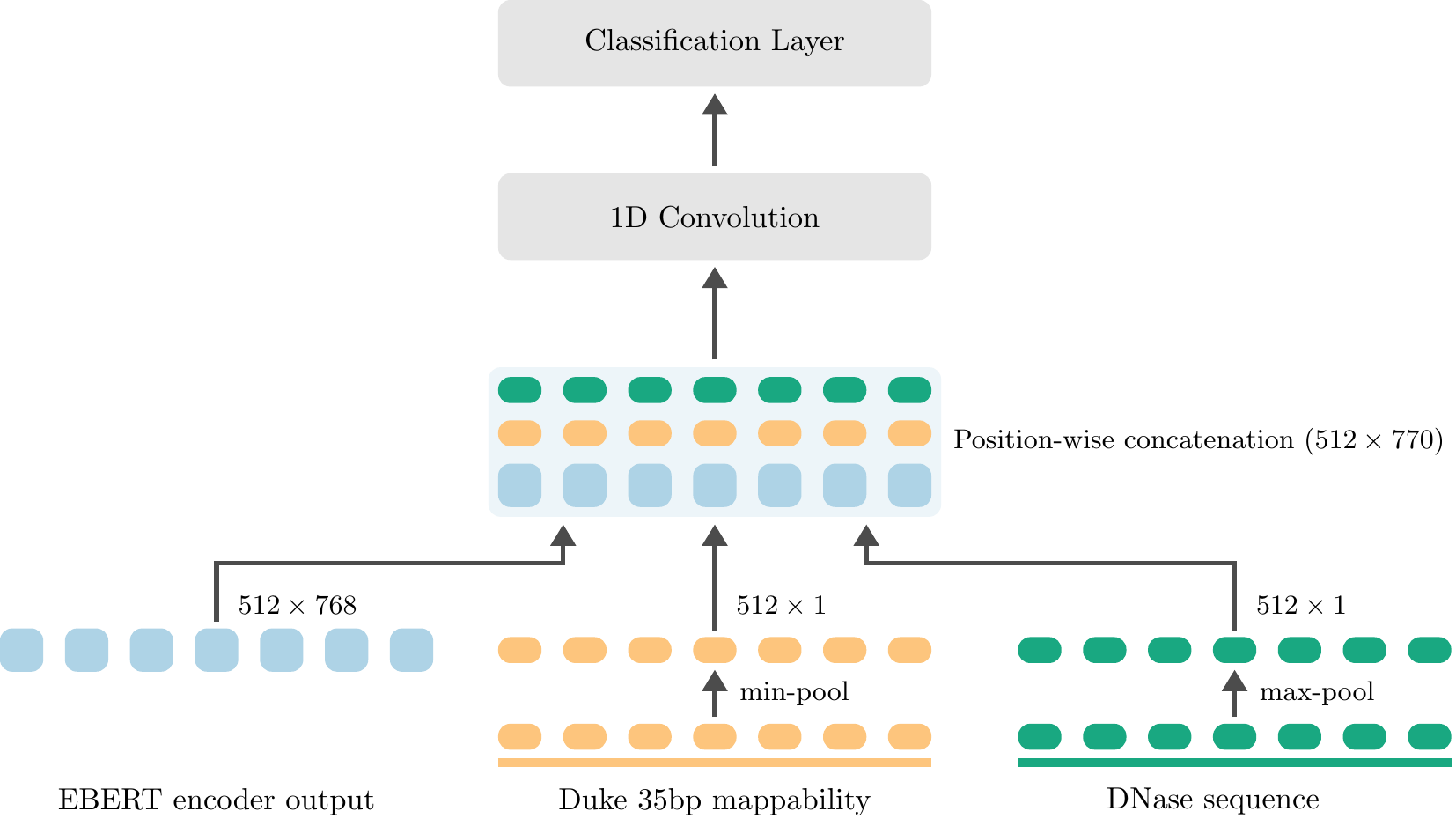}
  \caption{EBERT+ fine-tuning architecture. In the case of DBERT and EBERT fine-tuning, the BERT encoder output sequence of hidden states feeds directly into a 1D conv block with 2 1D convolutional layers followed by a 2-layer dense classification module. In the case of EBERT+, DNase-seq and mappability sequence features undergo pooling within sliding windows of size $k$ with stride $s$ to extract information matching the sequence resolution of the DNA/IDEAS tokens. These pooled auxiliary features are concatenated onto the sequence of hidden states from the BERT encoder, creating a final matrix of shape $L_{input} \times (H + 2)$. These concatenated outputs go through the convolutional and classification modules to produce the final prediction of TF binding.}
  \label{fig:arch2}
\end{figure}

\subsubsection{Performance Evaluation}

In addition to the top three models from the official ENCODE-DREAM final within-cell type benchmarking leaderboard, we include the recently published DeepGRN \citep{deepgrn} in our rankings, as this model reportedly achieved higher performance on several ENCODE-DREAM datasets. DeepGRN also exploits a Transformer architecture with DNA and DNase-seq inputs, which makes it an interesting comparison to our Transformer model with DNA, DNase-seq and IDEAS inputs.

The final rank on the ENCODE-DREAM leaderboard is an average rank across all 13 datasets. For each of the 13 datasets, a model’s rank was computed by averaging performance (on a set of held-out evaluation chromosomes) across 4 performance metrics: area under precision-recall curve (AUPRC), area under receiver-operator characteristic (AUROC), Recall at 50\% false discovery rate (FDR) and Recall at 10\% FDR. We will use these 4 metrics (on the evaluation set) to compare and contrast performance among our various models, and as absolute performance metrics with respect to the leaderboard. Given the extreme class imbalance of TF binding datasets, we will focus on AUPRC as the most informative metric with respect to true model performance.

\section{Results}

\subsection{Pre-training speedup on Cerebras Systems}

The training speedup afforded by the Cerebras system enabled us to explore architecture variations, tokenization schemes and hyperparameter settings in a way that would have been prohibitively time and resource intensive on a typical GPU cluster. DNABERT \citep{dnabert} reportedly required 25 days to pre-train on an 8-GPU cluster with effective batch size 2000. On a Cerebras CS-1 system, we pre-trained our DBERT model for twice the number of genome epochs in under 2 days with a batch size of 4096.  We pre-trained our EBERT model for ~1.75 epochs of 127 epigenomes in $\sim$2.5 days with batch size 8192, which we estimate would have taken $\sim$24 days of training on a GPU cluster with 16 nodes.

\subsection{Transcription factor binding prediction}

Our fine-tuning experiments show that both the addition of epigenetic data and the addition of auxiliary features each produce performance gains across all ENCODE-DREAM datasets. The final EBERT+ model, with both IDEAS state inputs and added TF binding specific auxiliary features, is competitive with the ENCODE-DREAM winners, beating all other models in 4/13 datasets and earning an overall rank of 3rd.

\begin{figure}[h]
  \centering
  \includegraphics[width=0.7\linewidth]{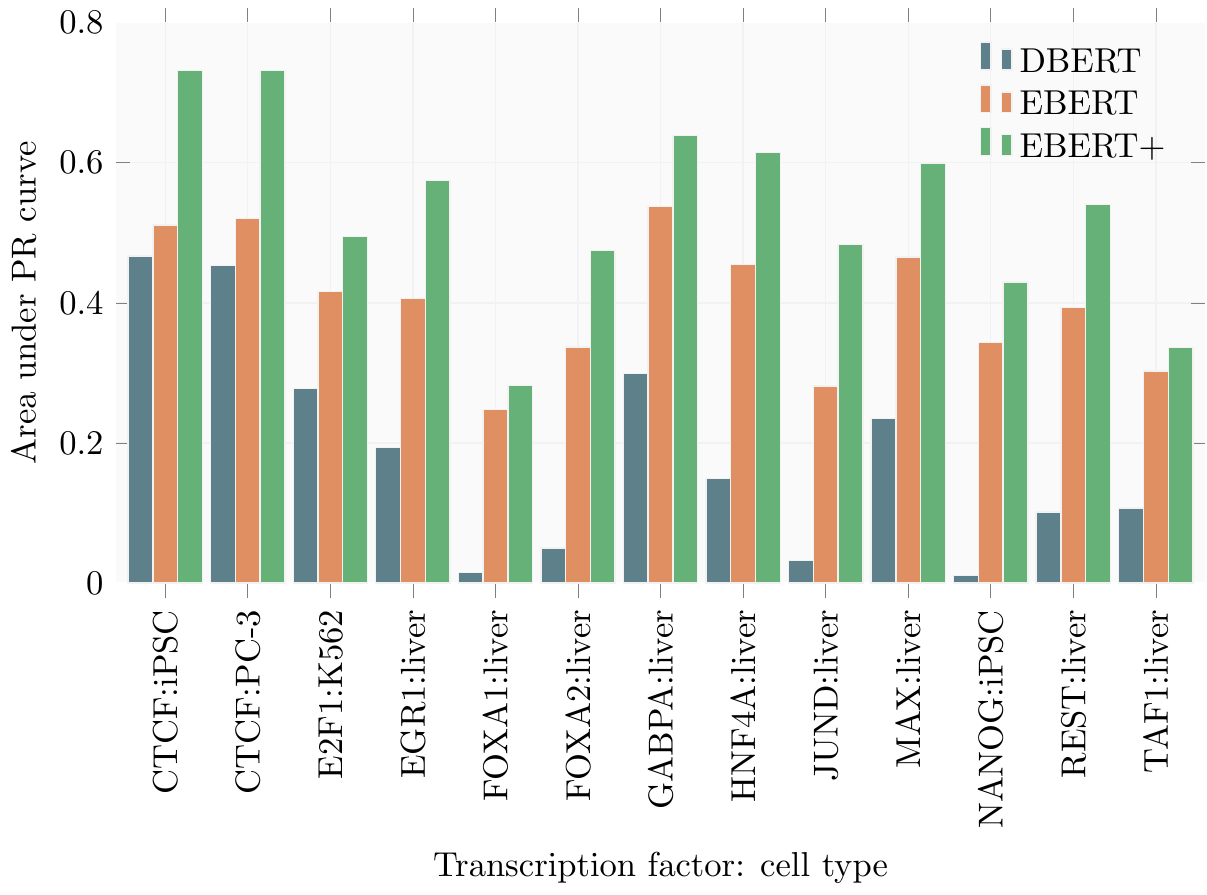}
  \caption{Cell type specific TF binding prediction performance as measured by AUPRC. For each of the 13 combinations of TF and cell type in the ENCODE-DREAM within-cell type benchmark data, we compute the AUPRC on the validation chromosomes for each fine-tuning strategy – DNA only (DBERT), DNA+IDEAS (EBERT), and DNA+IDEAS+auxiliary features (EBERT+). }
  \label{fig:all_auprc}
\end{figure}

Although the ENCODE-DREAM rankings use four performance metrics, we consider that the class imbalance in TF binding data make the AUPRC metric the most informative of the group. Figure \ref{fig:all_auprc} summarizes AUPRC values for our fine-tuning models across evaluation datasets.

We find EBERT+ has by far the highest (0.5405) median AUPRC across cell types, compared to EBERT (0.4061) and DBERT (0.1495), while EBERT and EBERT+ have similar variance across cell types (0.01 and 0.019 respectively).  In terms of the ENCODE-DREAM leaders, EBERT+ ranks 4th in terms of median AUPRC. Our overall leaderboard ranking of 3rd is attributable to higher performance across the full set of metrics.  

For CTCF datasets the largest performance boost comes from the addition of auxiliary features in EBERT+, with average increase of 0.21 AUPRC over EBERT (with average increase of just 0.06 AUPRC from DBERT to EBERT). For all other datasets, the largest increase in AUPRC is due to the addition of IDEAS state information in EBERT, with an average increase of 0.25 AUPRC over DBERT, versus an average 0.12 AUPRC increase from EBERT to EBERT+. 

\begin{figure}[h]
  \centering
  \includegraphics[width=1.0\linewidth]{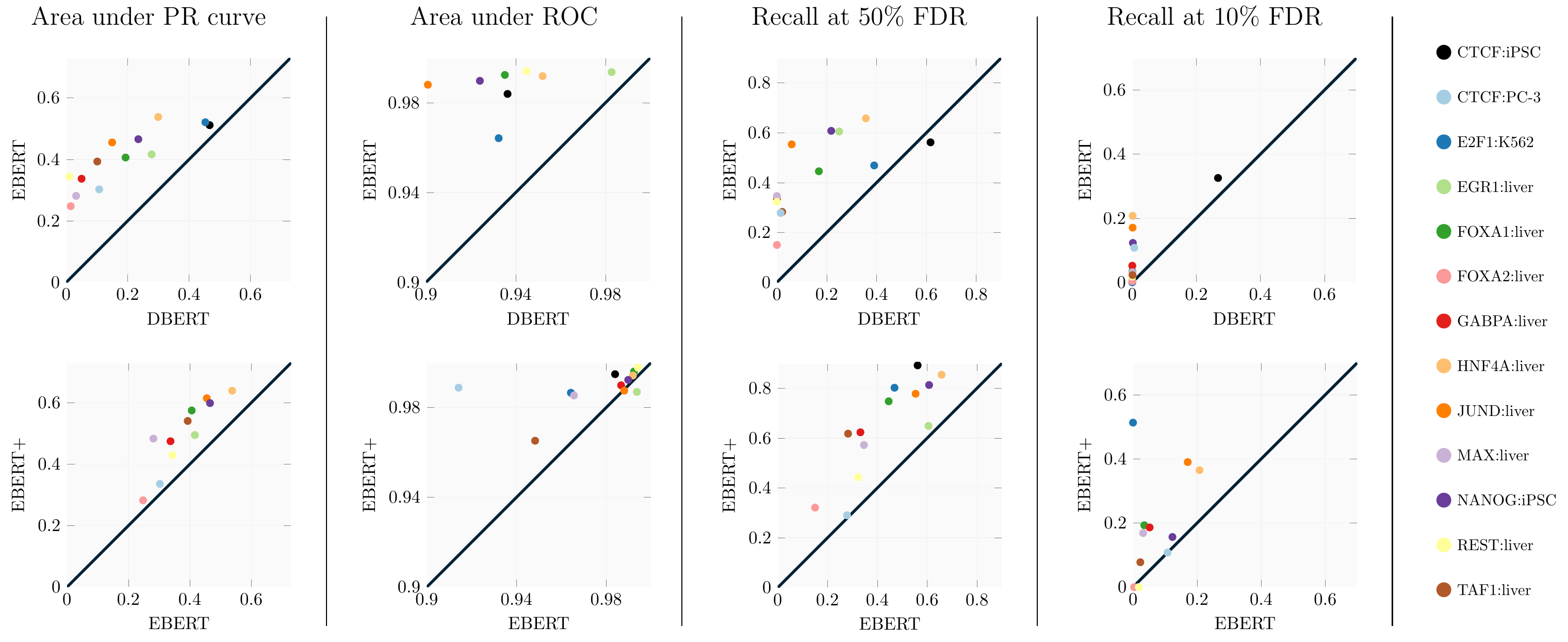}
  \caption{Top row: The effect of adding IDEAS state information at pre-training and fine-tuning. For each of the 13 test data sets, we compare the performance of DNA-only training (DBERT) and DNA+IDEAS training (EBERT) on each of the 4 ENCODE-DREAM ranking metrics. Data points falling in the top left triangle of each plot indicate superior performance due to inclusion of epigenetic inputs. Bottom row: The effect of adding DNase-seq and mappability auxiliary features to EBERT. For each of the 13 test data sets, we compare the performance of EBERT-only training (EBERT) and EBERT plus auxiliary features training (EBERT+) on each of the 4 ENCODE-DREAM ranking metrics. Data points falling in the top left triangle of each plot indicate performance gains due to inclusion of auxiliary features }
  \label{fig:comparison}
\end{figure}

To clarify how each TF-cell type dataset performance is affected by inclusion of epigenetic data during pre-training and fine-tuning, and inclusion of auxiliary features during fine-tuning, Figure \ref{fig:comparison} shows pairwise comparisons between DBERT and EBERT (epigenomic features), and between EBERT and EBERT+ (auxiliary features), for all four ENCODE-DREAM performance metrics (for underlying data see Table \ref{tab:tables1}).

\subsection{Impact of epigenetic feature: DBERT versus EBERT}

The addition of epigenomic features during pre-training and fine-tuning provides a performance boost in all metrics for the majority of ENCODE-DREAM datasets (12/13). GABPA in particular had a large performance boost across metrics, enough to put EBERT to 2nd place on the ENCODE-DREAM leaderboard for that dataset. Notably, both CTCF datasets had near identical fine-tuning performance from DBERT and EBERT models. CTCF binding is known to be dependent primarily on sequence motifs and is robust to chromatin state which is consistent with this result \citep{lambert}.

EBERT showed greater than 0.25 improvement in AUPRC for FOXA2, HNF4A, JUND and REST datasets, all in the liver cell type, with the largest improvement of 0.33 in the NANOG:iPSC dataset.

\subsection{Impact of auxiliary features: EBERT versus EBERT+}

In Figure \ref{fig:comparison} (leftmost panel) we see that the addition of DNase-seq and mappability features improves AUPRC between 0.033 and 0.22, with a median improvement of 0.14. This improvement is less than reported by other teams \citep{factornet, anchor, jteam}, potentially because the IDEAS state information included in EBERT already contains signal for chromatin accessibility across cell types.  

EBERT+ showed the largest improvement in AUPRC for CTCF and JUND datasets. 

\subsection{ENCODE-DREAM leaderboard}
The incorporation of IDEAS features was enough to place EBERT models into 2nd place for one dataset (GABPA:liver), and 4th place for 7 others, but was out of the top 5 for all other datasets. The auxiliary features included in EBERT+ further boosted performance on all datasets, leading to an overall 3rd place ranking among the 5 models compared. 

Table \ref{tab:table1} shows the per-dataset rankings of EBERT+ against other models, based on the combined evaluation set AUPRC, AUROC and Recall-at-xFDR metrics defined by the contest.  

EBERT+ had top overall performance on the EGR1, GABPA, HNF4A and MAX:liver datasets, with 12/13 datasets overall placing in the top 4 of 21 teams on the original leaderboard, and among the 5 models compared here. Interestingly, EBERT+ had better performance on the liver datasets compared to other cell types (enough to rank 2nd overall on liver datasets) and those datasets also showed the largest improvements due to IDEAS state rather than DNase. Liver was chosen as the cell type for the majority of the ENCODE-DREAM benchmark datasets in part because it is expected to be a difficult prediction task, as liver is not a single cell type but an amalgamation of several different types. It is possible the EBERT model performs well in this circumstance as it is not subject to measurement noise from multiple cell types as the direct-measured DNase-seq inputs.

\subsection{Impact of pre-training}

The EBERT+ model architecture performs less well if trained from randomly initialized weights, rather than from our pre-trained EBERT weights (Figure \ref{fig:all_auprc_supp}). It would still rank 1st in the GABPA:liver dataset, and 4th place overall. Our fine-tuned EBERT+ models outperformed the corresponding random-initialized model on AUPRC by 0.05-0.204 at convergence, and always converged after fewer training steps. DBERT and EBERT both also consistently outperform their random-initialized versions in this TF binding task.

\begin{table}[h!]
\caption{ENCODE-DREAM within-cell type benchmark dataset rankings}
\label{tab:table1}
\vskip 0.15in
\begin{center}
\begin{small}
\begin{tabular}{lccccc}
\toprule
\textbf{Dataset} & \textbf{EBERT+} & \textbf{JTeam} & \textbf{Factornet} & \textbf{Anchor} & \textbf{DeepGRN}\\
\midrule
CTCF:PC3   & 4 & 3 & \textbf{1} & 4 & \underline{2}\\
CTCF:iPSC & 4 & 3 & \textbf{1} & 5 & \textbf{1}\\
E2F1:K562 & \underline{2} & \textbf{1} & 4 & 3 &5\\
EGR1:liver  & \textbf{1} & \underline{2} &4 & 3 &5\\ 
FOXA1:liver & 5 & 3 & \textbf{1} &4 &\underline{2}\\
FOXA2:liver & 3 & \underline{2} & \textbf{1} &5 &4\\
GABPA:liver & \textbf{1} & 3 & 4 &\underline{2} &5\\
HNF4A:liver & \textbf{1} & 4 & \underline{2} &3  &5\\
JUND:liver   & 4 & \underline{2} & 3 &\textbf{1} &4\\
MAX:liver     & \textbf{1} & 3 & 4 &\underline{2} &5\\
NANOG:iPSC&3 & \underline{2} & 4  &5 &\textbf{1}\\
REST:liver    & 4 & \textbf{1} & 3 &\underline{2} &4\\
TAF1:liver     & 4 & \textbf{1} & 3 &\underline{2} &5\\
\midrule\midrule
Overall       & 3 & \textbf{1} & \underline{2} & 4 & 5\\
\bottomrule
\end{tabular}
\end{small}
\end{center}
\vskip -0.2in
\end{table}


\section{Discussion}

We have introduced a novel class of genomic language model that learns embeddings informed by both upstream and downstream DNA sequence and surrounding cell type-specific epigenetic state context.

We exploited the Cerebras CS-1 system to conduct self-supervised pre-training runs of unprecedented scale in the ‘biological language’ domain. The BERT$_{BASE}$ architecture \citep{devlin} employed in this work has 12 layers with 12 attention heads apiece, for a total of $\sim$100M parameters. Our pre-training dataset contains 127 paired epigenomes. With a human genome size of 3 billion bp -- 3 million samples after tokenization -- our dataset is equivalent to $\sim$380M paired samples for pre-training. Given the complexity of epigenomic structures, and the size of our dataset, we hypothesize that an EBERT model with larger capacity may produce better performance on downstream tasks, as evidence from Transformer-based language models shows that optimal performance is achieved when dataset size and model size scale together \citep{kaplan}.

Tests on the upcoming CS-2 system -- which features more than twice the compute and memory of the CS-1 -- demonstrate an increase in pre-training throughput of $\sim$2$\times$ for EBERT$_{BASE}$. The CS-2 will further shorten training times for larger EBERT models and will make it even easier to expand beyond the BERT$_{BASE}$ architecture; EBERT$_{LARGE}$ (24 $\times$ 16 $\times$ 1024) pre-training on CS-2 runs at approximately the same throughput as EBERT$_{BASE}$ on the CS-1.

To begin to evaluate the application of EBERT epigenomic language models, we fine-tuned them to perform a challenging but critical task in regulatory genome analysis, that of cell type specific TF binding prediction. These models were competitive with leading models in the literature and provide evidence that the addition of epigenetic information at the pre-training stage improved performance on downstream tasks in which chromatin context interpretation is a key element.

The key feature that makes EBERT unique with respect to our ENCODE-DREAM competitors is generality. The other leaderboard winners are all explicitly and exclusively TF binding models. While we chose TF binding as an illustrative task here, the EBERT model can also be easily adapted to fine-tune any number of other cell type specific classification tasks – gene expression, splicing, or 3D-interactions to name a few possibilities.

Future work will explore both increasing the capacity of pre-trained EBERT models and expanding the range of fine-tuning applications to which EBERT is applied.

\section{Conclusion}
We trained our epigenomic language model EBERT on a very large dataset of 127 paired epigenomes. We used EBERT to train fine-tuned TF binding models that are competitive with leading models in the literature and showed the addition of epigenetic information at the pre-training stage improves performance. Including epigenetic data alongside DNA sequence data in a language model may offer major benefits to a variety of sequence based genomic classification tasks. Training these complex models has previously been a computationally intractable problem. Our partnership with Cerebras enables us to rapidly explore the benefits of using tissue specific datasets in our models. Future work will include further developments of the proposed architecture, the use of larger, more accurate models, enabled by the performance of the Cerebras CS-2 system, greater quantities and quality of epigenetic data, and tests on a wider variety of datasets and fine-tuning tasks.

\section{Acknowledgements}
The authors thank Petr Votava, Pavel Kovtunenko and Ross Walker for hardware support; Eugene Vecharynski, Abhay Gupta, Tanveer Raza for assistance with Cerebras systems; Natalia Vassilieva, Rebecca Lewington, Paul Smyth for comments on the manuscript; Francesco Farina for assistance with data visualization and Laura Acqualagna for contributions to early stages of the project.

\bibliographystyle{abbrvnat}
\bibliography{biblio}

\pagebreak
\begin{center}
\textbf{\Large Supplementary Materials}
\end{center}
\setcounter{equation}{0}
\setcounter{figure}{0}
\setcounter{table}{0}
\setcounter{section}{0}
\setcounter{page}{1}
\makeatletter
\renewcommand{\theequation}{S\arabic{equation}}
\renewcommand{\thefigure}{S\arabic{figure}}
\renewcommand{\thetable}{S\arabic{table}}
\renewcommand{\bibnumfmt}[1]{[S#1]}
\renewcommand{\citenumfont}[1]{S#1}
%
\section{Data}
For DNA sequence input we use the reference human genome version hg19/GRCh37. The IDEAS state dataset (\url{http://bx.psu.edu/~yuzhang/ENCODE-36state/}) identifies 36 IDEAS states at 200bp resolution, in 127 different tissue types.

\section{Model training}

All models were implemented in Python using the Estimator API of TensorFlow 2.0.

\subsection{Pre-training}

All pre-training experiments were conducted on a Cerebras CS-1 system using models compiled with Cerebras Graph Compiler. The EBERT$_{BASE}$ model was trained on batches of 8,192 sequences of 1,000bp for a total of 70,000 batches at time of loss convergence. This is approximately equivalent to 1.75 epochs of training on 127 epigenomes. This training procedure required $\sim$2.5 days, or 1.5 days per 127-epigenome epoch. Our DBERT model was pre-trained on batches of 4,096 sequences of 1,000bp for 120,000 steps, taking $\sim$40 hours to complete >100 genome epochs. In both models chromosomes 8 and 21 were held out for evaluation, with all other chromosomes used for training.
 
\subsection{Fine-tuning}

We used the within-cell benchmarking datasets from the ENCODE-DREAM Challenge as a base for our experiments. There are 13 TF-cell type datasets, for which we used DNA sequence, DNase-seq and Duke 35bp mappability uniqueness data, and supplemented this with IDEAS state data. Processed data files for the DNase-seq and mappability features are available from the FactorNet Repository (\url{https://github.com/uci-cbcl/FactorNet/tree/master/resources}).

After tokenization into $k$-mers to match the DNA and IDEAS tokens, we applied max-pooling to each DNase-seq $k$-mer to increase sensitivity to peaks, and min-pooling to mappability $k$-mers so that any low-uniqueness signal produces a low uniqueness token. Following ENCODE-DREAM rules, chromosomes 2-7, 9-20, 22 and X were used for model training, with chromosomes 1, 8 and 21 held out for evaluation. The genome was segmented into overlapping 200bp bins with a 50bp sliding window. Each bin was labeled as Bound, Unbound or Ambiguous depending on the majority label of the 200bp bin. Bins labeled as Ambiguous were excluded from evaluation. For training, we took each 200bp bin and expanded the sequence 400bp on each side to attain a starting sequence of 1,000bp per sample. We use this 1,000bp window to produce associated IDEAS, DNase-seq and mappability sequence features. For a more complete description of the datasets please see the ENCODE-DREAM Challenge website.

We trained each fine-tuning model using datasets with enforced 10:1 negative: positive examples, with negative examples resampled after each epoch. Here an epoch means having seen the full set of positive examples once. Models were trained on 16 GPUs with a maximum learning rate of 0.0001 after a a linear warmup of 25,000 steps. Training was halted when evaluation AUPRC plateaued. See Table \ref{tab:tables2} for full hyperparameters. 

\pagebreak

\begin{table}[]
\caption{ENCODE-DREAM within-cell type benchmark performance metrics}
\label{tab:tables1}
\vskip 0.15in
\begin{center}
\begin{small}
\begin{tabular}{llcccc}
\toprule
\textbf{Dataset} & \textbf{Model}  & \textbf{AUPRC} & \textbf{AUROC} & \textbf{Re@10FDR} & \textbf{Re@50FDR}\\
\midrule
\multirow{3}{*}{CTCF:iPSC} & \textbf{DBERT} & 0.4669 & 0.9364 & 0.2676 & 0.6164\\
& \textbf{EBERT} &  0.5113	& 0.9840	 & 0.3256	 & 0.5616\\
&\textbf{EBERT+} & 0.7314 & 0.9949 & 0.7021 & 0.8917\\
\midrule
\multirow{3}{*}{CTCF:PC-3 } & \textbf{DBERT} & 0.4592 & 0.9324 & 0.0000 & 0.3903\\
& \textbf{EBERT} & 0.5209 & 0.9642 & 0.0000 & 0.4689 \\
&\textbf{EBERT+} & 0.7316 & 0.9866 & 0.5131 & 0.8020\\
\midrule
\multirow{3}{*}{E2F1:K562} & \textbf{DBERT} & 0.2780 & 0.9827 & 0.0000 & 0.2500\\
& \textbf{EBERT} &  0.4164 & 0.9937 & 0.0046 & 0.6051 \\
&\textbf{EBERT+} &  0.4948 & 0.9870 & 0.0000 & 0.6097\\
\midrule
\multirow{3}{*}{EGR1:liver} & \textbf{DBERT} & 0.1932 & 0.9351 &0.0000&0.1683\\
& \textbf{EBERT} &  0.4061 & 0.9924 & 0.0350 & 0.4455\\
&\textbf{EBERT+} & 0.5747 & 0.9960 & 0.1931 & 0.7475\\
\midrule
\multirow{3}{*}{FOXA1:liver} & \textbf{DBERT} &0.0147 & 0.7927 &0.0000&0.0000\\
& \textbf{EBERT} &  0.2478 & 0.8253 & 0.0033 & 0.1500\\
&\textbf{EBERT+} & 0.2825 & 0.8772 & 0.0000 & 0.3211\\
\midrule
\multirow{3}{*}{FOXA2:liver} & \textbf{DBERT} &0.0501 & 0.8911 &0.0000&0.0000\\
& \textbf{EBERT} &  0.3371 & 0.9866 & 0.0519 & 0.3321\\
&\textbf{EBERT+} & 0.4746 & 0.9899 & 0.1862 & 0.6232\\
\midrule
\multirow{3}{*}{GABPA:liver} & \textbf{DBERT} & 0.2995 & 0.9520 & 0.0011 & 0.3570\\
& \textbf{EBERT} & 0.5379 & 0.9919 & 0.2076 & 0.6577\\
&\textbf{EBERT+} & 0.6388 & 0.9943 & 0.3647 & 0.8541\\
\midrule
\multirow{3}{*}{HNF4A:liver} & \textbf{DBERT} & 0.1495 & 0.9008 & 0.0011 & 0.0593\\
& \textbf{EBERT} &  0.4552 & 0.9881 & 0.1707 & 0.5534 \\
&\textbf{EBERT+} & 0.6143 & 0.9875 & 0.3900 & 0.7778\\
\midrule
\multirow{3}{*}{JUND:liver } & \textbf{DBERT} & 0.0323 & 0.8256 &0.0000&0.0000\\
& \textbf{EBERT} &  0.2814 & 0.9656 & 0.0315 & 0.3462\\
&\textbf{EBERT+} & 0.4830 & 0.9854 & 0.1685 & 0.5722\\
\midrule
\multirow{3}{*}{MAX:liver} & \textbf{DBERT} & 0.2347 & 0.9240 & 0.0018 & 0.2179\\
& \textbf{EBERT} &  0.4657 & 0.9898 & 0.1223 & 0.6076\\
&\textbf{EBERT+} & 0.5986 & 0.9923 & 0.1564 & 0.8128\\
\midrule
\multirow{3}{*}{NANOG:iPSC} & \textbf{DBERT} & 0.0108 & 0.9449 &0.0000&0.0000\\
& \textbf{EBERT} &  0.3438 & 0.9941 & 0.0180 & 0.3234\\
&\textbf{EBERT+} & 0.4295 & 0.9978 & 0.0000 & 0.4431\\
\midrule
\multirow{3}{*}{REST:liver} & \textbf{DBERT} & 0.1013 & 0.8222 &0.0008&0.0212\\
& \textbf{EBERT} &  0.3931 & 0.9483 & 0.0228 & 0.2826\\
&\textbf{EBERT+} & 0.5405 & 0.9652 & 0.0778 & 0.6177\\
\midrule
\multirow{3}{*}{TAF1:liver} & \textbf{DBERT} & 0.1074 & 0.8695 &0.0059&0.0150\\
& \textbf{EBERT} &  0.3025 & 0.9142 & 0.1078 & 0.2784 \\
&\textbf{EBERT+} & 0.3359 & 0.9889 & 0.1078 & 0.2904\\
\bottomrule
\end{tabular}
\end{small}
\end{center}
\vskip -0.2in
\end{table}

\subsection{Effect of pre-training}

\begin{figure}[h!]
  \centering
  \includegraphics[width=0.8\linewidth]{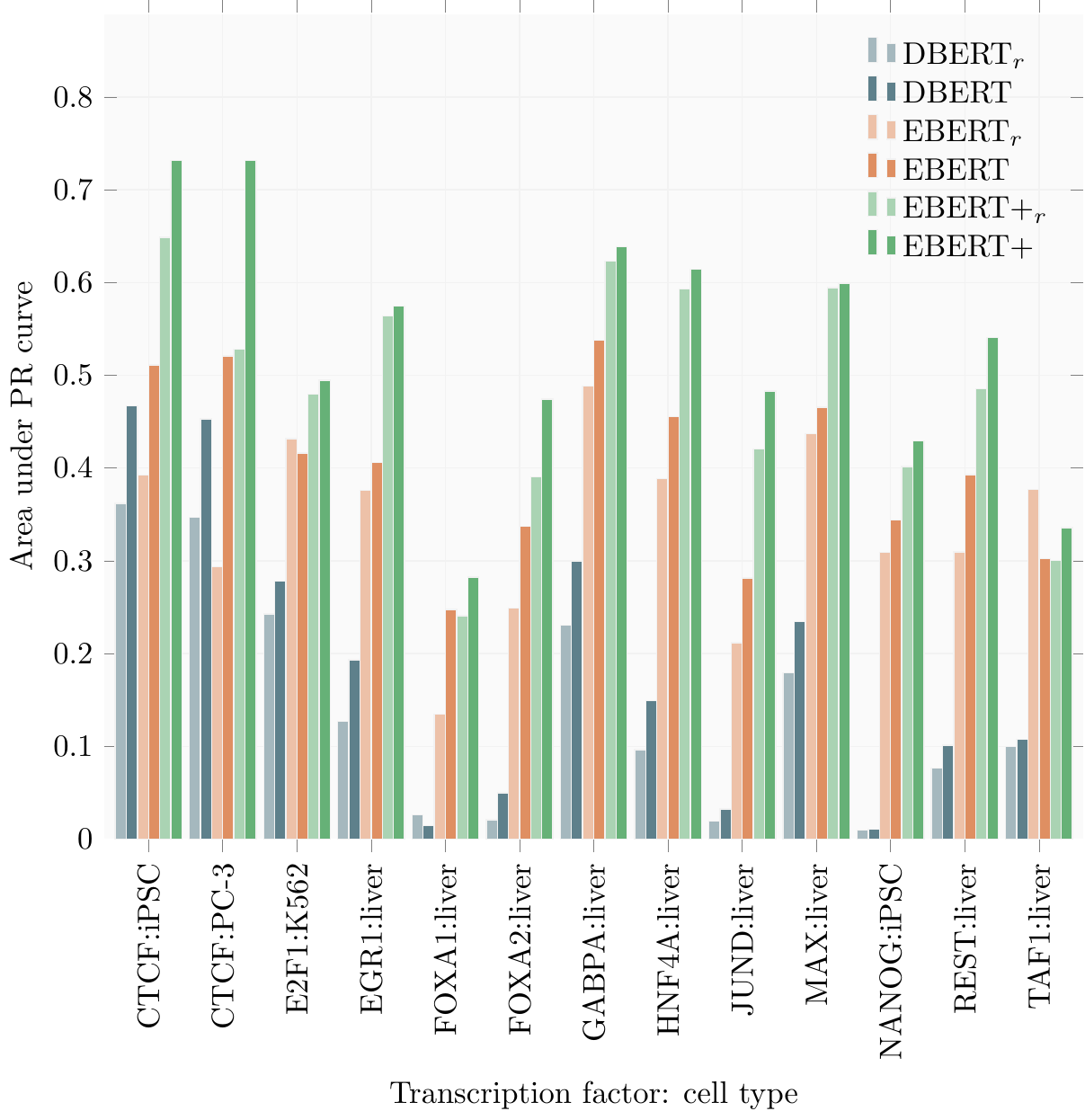}
  \caption{Cell type-specific TF binding prediction performance for randomly initialized versus pre-trained models. For each of the ENCODE-DREAM within-cell type benchmark datasets, we compute the AUPRC on the validation chromosomes for each fine-tuning strategy – DNA only (DBERT), DNA+IDEAS (EBERT), and DNA+IDEAS+auxiliary features (EBERT+). For each architecture, we show a model$_r$ trained from randomly initialized weights.}
  \label{fig:all_auprc_supp}
\end{figure}

\pagebreak

\subsection{Hyperparameters}
The CS-1 system allowed us to experiment with many different EBERT configurations, with the best used as bases for fine-tuning models. The hyperparameters used for models reported in this paper are shown in Table \ref{tab:tables2}.

\begin{table}[h!]
\caption{Model training hyperparameters}
\label{tab:tables2}
\vskip 0.15in
\begin{center}
\begin{small}
\begin{tabular}{lccc}
\toprule
\textbf{Hyperparameter} & \textbf{DBERT} & \textbf{EBERT} & \textbf{TF binding} \\
\midrule
      Input size (bp) & 1000 & 1000 & 1000 \\
      Input sequence length & 150 & 150 & 150 \\
      $k$-mer size & 7 & 7 & 7\\
      Tokenization stride & 7 & 7&7 \\
      Peak learning rate & 0.0004 & 0.00006 & 0.0001\\
      Warmup steps & 24000 & 30000 & 25000\\
      Batch size & 4096 & 8192 & 320\\
      Training class balance & N/A & N/A & 10:1 \\
      Convolution layers & & & 2\\
      Dense layers & & & 2 \\
      Optimizer & AdamW & AdamW & AdamW \\
      Adam $\beta1$|$\beta2$|$\epsilon$ & 0.9$|$0.98$|$0.0001 & 0.9$|$0.98$|$0.0001 & 0.9$|$0.99$|$0.00001\\
      Loss scale & 16000 & 16000 & 16000\\
      Attention dropout & 0.1 & 0.1 & \\
      Dropout & 0.1 & 0.1 &\\
      Attention heads & 12 & 12 & 12 \\
      Layers & 12 & 12 & 12\\
      Filter size & 3072 & 3072 & 3072\\
      Hidden size & 768 & 768 & 768\\
\bottomrule
\end{tabular}
\end{small}
\end{center}
\vskip -0.2in
\end{table}
%
%

\end{document}